\documentclass[preprint,12pt, a4paper]{elsarticle}

\usepackage{amssymb}
\usepackage[breaklinks=true]{hyperref}
\usepackage{algorithm}
\usepackage{graphicx}
\usepackage[noend]{algpseudocode}

\usepackage{array}
\usepackage{booktabs}
\usepackage[usenames,dvipsnames]{color}
\usepackage{enumitem}
\usepackage{caption}
\usepackage{adjustbox}
\usepackage{setspace}
\usepackage[section]{placeins}
\usepackage{gensymb}
\usepackage{mathtools}
\usepackage{amsthm}
\usepackage{lineno}
\usepackage{rotating}
\usepackage{caption}
\usepackage{listings}
\usepackage{subcaption}

\usepackage{tabularx}
\usepackage{multirow}

\usepackage[OT4]{fontenc}
\usepackage[utf8]{inputenc}

\journal{ }

\newcommand{\q}{\phantom{0}}

\newcommand{\link}[1]{\url{#1}}

\lstdefinelanguage{Python}
{
	basicstyle=\fontsize{6}{7}\selectfont\ttfamily,
	keywords={import, as, from},
	morekeywords=[2]{axis, minsupp_new, induction_measure, pruning_measure, voting_measure, labels, normalize, cmap},
	keywordstyle=[1]\color{blue},
	keywordstyle=[2]\color{Orange},
	comment=[l]{\#},
	commentstyle=\color{darkgreen},
	string=[b]{'},
	stringstyle=\color{Violet},
	tabsize=3,
	belowskip=0cm
}

\begin{document}
\renewcommand{\labelenumii}{\arabic{enumi}.\arabic{enumii}}

\begin{frontmatter}



\title{RuleKit 2: Faster and simpler rule learning}


\author[polsl,emag]{Adam Gudy\'s\corref{mycorrespondingauthor}}
\ead{adam.gudys@polsl.pl}
\author[polsl,emag]{Cezary Maszczyk}
\author[emag]{Joanna Badura}
\author[emag]{Adam Grzelak}
\author[polsl,emag]{Marek Sikora}
\author[polsl,emag]{\L{}ukasz Wr\'obel}

\cortext[mycorrespondingauthor]{Corresponding author}
\address[polsl]{Faculty of Automatic Control, Electronics and Computer Science, Silesian University of Technology, Akademicka 16, 44-100 Gliwice, Poland}
\address[emag]{Łukasiewicz Research Network –- Institute of Innovative Technologies EMAG, Leopolda 31, 40-189 Katowice, Poland}

\begin{abstract}

Rules offer an invaluable combination of predictive and descriptive capabilities. Our package for rule-based data analysis, RuleKit, has proven its effectiveness in classification, regression, and survival problems. Here we present its second version. New algorithms and optimized implementations of those previously included, significantly improved the computational performance of our suite, reducing the analysis time of some data sets by two orders of magnitude. The usability of RuleKit 2 is provided by two new components: Python package and browser application with a graphical user interface. The former complies with \textit{scikit-learn}, the most popular data mining library for Python, allowing RuleKit 2 to be straightforwardly integrated into existing data analysis pipelines. RuleKit 2 is available at GitHub under GNU AGPL~3 license (\link{https://github.com/adaa-polsl/RuleKit}). 


\end{abstract}

\begin{keyword}
decision rules \sep classification \sep regression \sep survival analysis \sep knowledge discovery \sep explainable artificial intelligence \sep Python package \sep graphical user interface



\end{keyword}

\end{frontmatter}

\newpage

\section*{Metadata}
\label{}

\begin{table}[!h]
\begin{tabular}{|l|p{6.5cm}|p{6.5cm}|}
\hline
\textbf{Nr.} & \textbf{Code metadata description} & \textbf{Metadata} \\
\hline
C1 & Current code version & v2.1.24 \\
\hline
C2 & Permanent link to code/repository used for this code version & \url{https://github.com/adaa-polsl/RuleKit/releases/tag/v2.1.24} \\
\hline
C3  & Permanent link to Reproducible Capsule & \url{https://adaa-polsl.github.io/RuleKit-python/serve/v2.1.24.1/rst/tutorials.html}\\
\hline
C4 & Legal Code License   & GNU AGPL v3.0 \\
\hline
C5 & Code versioning system used & git \\
\hline
C6 & Software code languages, tools, and services used & Java, Python, Streamlit \\
\hline
C7 & Compilation requirements, operating environments \& dependencies & Java JDK 11\\
\hline
C8 & If available Link to developer documentation/manual & Manuals: \url{https://github.com/adaa-polsl/RuleKit/wiki} \\
& & \url{https://adaa-polsl.github.io/RuleKit-python/serve/index.html}\\
& & JavaDoc: \url{https://adaa-polsl.github.io/RuleKit/}\\
\hline
C9 & Support email for questions & adam.gudys@polsl.pl \\
\hline
\end{tabular}
\caption{Code metadata}
\label{codeMetadata} 
\end{table}



\section{Motivation and significance}

Thanks to their interpretability, symbolic models are indispensable for decision-making in critical applications where prediction process has to be verified by an expert (e.g., in medicine). This area of machine learning is referred to as explainable artificial intelligence (XAI) and, with a growing use of AI systems in various human activities, becomes of increasing importance as well~\citep{xai2016,Arrieta2020}. Among symbolic models, also known as white boxes, decision rules are the easiest to understand by a human and also the most general. This is because rules, unlike branches of decision trees, may overlap and can be considered independently as local explanations of a data set~\citep{Lakkaraju2016}. This feature plays crucial role in XAI systems, e.g., LIME~\citep{Ribeiro2016} or LUX~\citep{Bobek2025}, where rules are used to explain decision of complex, non-interpretable models. 
However, the locality feature is important not only for descriptive purposes -- it is employed by rule-based ensembles like ENDER~\citep{dembczynski2010} or BOOMER~\citep{Rapp2020} where high predictive capabilities are obtained by learning many overlapping rules in ``problematic'' areas of an attribute space.    

RuleKit~\cite{gudys2020rulekit}, our package for rule-based learning in classification, regression, and survival data has been successfully applied in a number of critical areas. These are industrial safety (forecasting high energy seismic bumps and predicting methane concentration in coal mines~\citep{sikora2019}), medicine (survival prognosis in breast, lung, thyroid and blood cancers~\citep{wrobel2017,Kulis2022}), or the analysis of multi-omics data~\citep{Gruca2021}. The performance of the first RuleKit version was thoroughly compared with competing packages like Orange~\citep{demvsar2013orange}, Rseslib~\citep{rseslib2019}, or Weka~\citep{Witten2016}. While many other tools provides algorithms for rule-based learning, most of them concerns association rules (e.g., arules~\citep{Hahsler2005}, SPMF~\citep{fournier2016spmf}, mlxtend~\citep{Raschkas_2018_mlxtend}, EasyMiner~\citep{Vojivr2020}, or PAMI~\citep{Kiran_2025_PAMI}), thus cannot be directly compared with RuleKit.

Here, we present RuleKit 2. New algorithms and optimized implementations of those previously included significantly improved the scalability of our suite, reducing analysis time of some data sets by two orders of magnitude. This comes with an extended usability provided by two new components: Python package and browser application with a graphical user interface.


\section{Software description}


RuleKit 2 is a versatile package for rule-based learning suitable for classification, regression, and survival data. As in the predecessor, the learning algorithm follows separate and conquer (a.k.a, sequential covering) paradigm~\citep{furnkranz1999} with individual rules being found using a greedy hill-climbing heuristic~\citep{furnkranz2012foundations} driven by a selected quality measure~\citep{janssen2010quest}.

\subsection{Software architecture}

\begin{figure*}[t!]
	\centering
	\includegraphics[width=\textwidth]{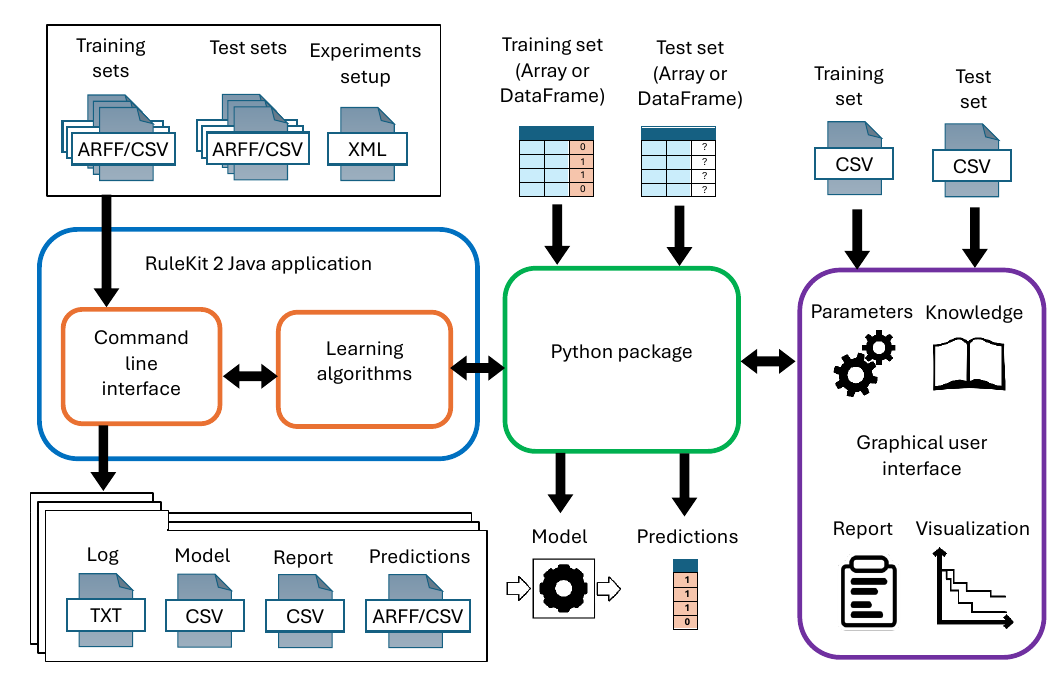}	
	\caption{RuleKit 2 architecture.}\label{fig:architecture}
\end{figure*}

RuleKit 2 is composed of three components: a command line tool, a Python package, and a browser application with a graphical user interface (see Figure~\ref{fig:architecture} for the architecture overview). Unlike RuleKit 1, the second revision does not depend on RapidMiner~\citep{rapidminer}.

The first module, implemented as a standalone Java application, integrates all RuleKit 2 algorithms and provides user with an experimental environment based on XML. It allows batch analyses of many datasets using various algorithm configurations, generating comprehensive reports. Compared to RuleKit 1, it provides new algorithms and optimized, more configurable implementations of those previously included. This is facilitated with an improved versatility and flexibility of the testing environment. 

The second component, a RuleKit-python package, allows performing rule-based analyses in Python. By directly accessing RuleKit 2 Java classes, Python package introduces no overhead compared to using XML-based batch interface, at the same time being more flexible. This is a large advantage over R package included in RuleKit 1, which prepared XML files and invoked a RuleKit command line tool underneath.              

Finally, Rulekit 2 provides a browser application wrapping the Python package in a friendly graphical user interface. This module was implemented in Streamlit and replaces the RapidMiner plug-in from RuleKit 1.

 \subsection{Software functionalities}

RuleKit 2 provides multiple novelties in all its components.  

\vspace{0.3cm}
\textbf{Java application}. The updates in this module include mostly algorithmic changes, for instance: 
\begin{itemize}[noitemsep,topsep=0pt,parsep=0pt,partopsep=0pt]
	\item{more flexible user-guided induction (possibility to enforce a desired number of rules; adjustment of conditions)},
	\item{integrated contrast sets mining with RuleKit-CS algorithm~\citep{Gudys2024},}
	\item{time and memory optimization of the existing induction algorithms (improved computational performance of hypergeometric test, log-rank test, Kaplan-Maier estimator, and some other procedures),}
	\item{new algorithm for inducing regression rules based on mean faster by $O(|D|)$ factor than the previous median-based variant, with $|D|$ being the number of examples.} 
\end{itemize}

\vspace{0.3cm}
\textbf{RuleKit-python package}. 
This component replaces R package from the first version of RuleKit, offering significantly broader functionality. As the interface of RuleKit-python complies with \textit{scikit-learn}, the most popular data mining library for Python, RuleKit~2 can be straightforwardly integrated into existing analysis pipelines as an alternative to other machine learning methods. Data sets can be provided as array-like Python structures, thus the input/output is not limited to ARFF/CSV files as in the command line tool, but includes all Python-readable formats (e.g., XLS/XLSX, HTML, XML, JSON, SQL databases).

\vspace{0.3cm}
\textbf{Graphical user interface}.
In contrast to RuleKit 1, where the graphical user interface was provided by the RapidMiner plug-in, the second revision includes a browser application wrapping RuleKit-python. This made configuration of algorithms' parameters and defining expert knowledge more intuitive. New functionalities like visualization of survival function estimates were also added.
  

\section{Illustrative examples}

\begin{table}
	\centering
	\begin{scriptsize}
		\setlength{\tabcolsep}{0.33em}
		\begin{tabular}{lp{0.27em}cccp{0.27em}cccp{0.27em}ccc}
			&& \multicolumn{3}{c}{Classification (50 data sets)} && \multicolumn{3}{c}{Regression (48 data sets)} && \multicolumn{3}{c}{Survival (22 data sets)}\\
			\cline{3-5}  \cline{7-9} \cline{11-13} 
			&&  \#Rules & BAcc & Time &&  \#Rules & RRSE & Time && \#Rules & IBS & Time \\
			&& 	(avg)	& (avg)	& (sum)	&& 	(avg)	& (avg)		& (sum) && 	(avg)	& (avg)		& (sum)	\\
			\hline
			RuleKit 1 && 45.0 & 0.779 & 33m\,38s && 25.2 & 0.898 & 12h\,53m\,47s && 9.2 & 0.155 & 9h\,26m\,00s \\
			RuleKit 2 && 48.8 & 0.782 & \q4m\,07s && 25.8 &  0.722 & \q2h\,12m\,14s && 9.8 & 0.155 & 2h\,35m\,20s \\
			
		\end{tabular}
	\end{scriptsize}
	\caption{Results of RuleKit 1 and RuleKit 2 averaged/summed over selected data sets (each run in 10-fold cross validation): number of rules (\#Rules), balanced accuracy (BAcc), root relative squared error (RRSE), integrated Brier score (IBS). The algorithms were run with default parameters.}
	\label{tab:results}
\end{table}

\begin{figure}
	\includegraphics[width=1.0\textwidth]{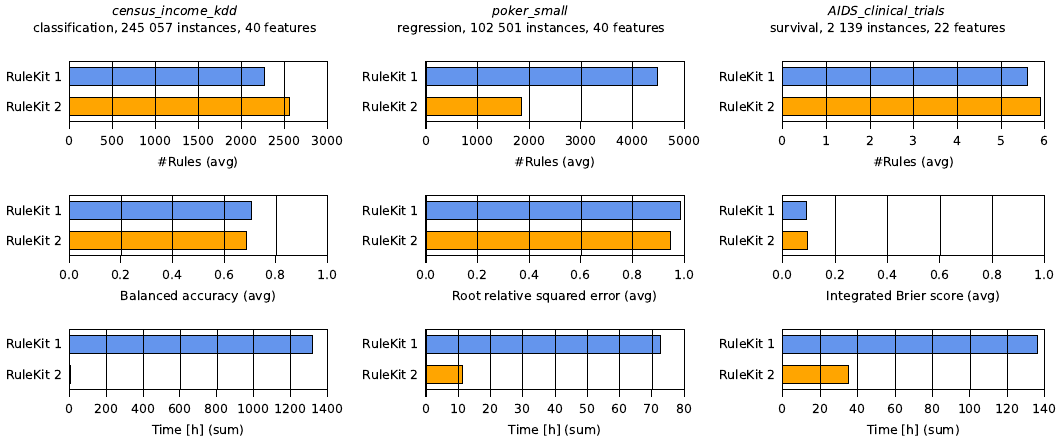}
	\caption{RuleKit performance in three selected large data sets (classification, regression, and survival). The experiments were performed in 10-fold cross validation.}\label{fig:big-data}
\end{figure}

\begin{figure}[t]
	\begin{minipage}{7.5cm}
		\begin{lstlisting}[language=Python] 
			import pandas
			import matplotlib.pyplot as plt
			from sklearn.metrics import ConfusionMatrixDisplay
			from rulekit.classification import RuleClassifier
			from rulekit.params import Measures
			
			URL = 'https://github.com/adaa-polsl/RuleKit/' 
				+ 'raw/refs/heads/master'
			
			train = pandas.read_parquet(URL + '/data/car/train.parquet')
			X, y = train.drop('class', axis=1), train['class']
			
			test = pandas.read_parquet(URL + '/data/car/test.parquet')
			X_test, y_test = test.drop('class', axis=1), test['class']
			
			clf = RuleClassifier(
				minsupp_new=1, 
				induction_measure=Measures.C2,
				pruning_measure=Measures.C2, 
				voting_measure=Measures.Correlation)
				clf.fit(X, y)
			
			disp = ConfusionMatrixDisplay.from_predictions(
				y_test, clf.predict(X_test), labels=y.unique(), 
				normalize='pred', cmap='Blues')
		\end{lstlisting}
		\subcaption{}
		\centering
		\vspace{0.3cm}
		\includegraphics[width=0.75\textwidth]{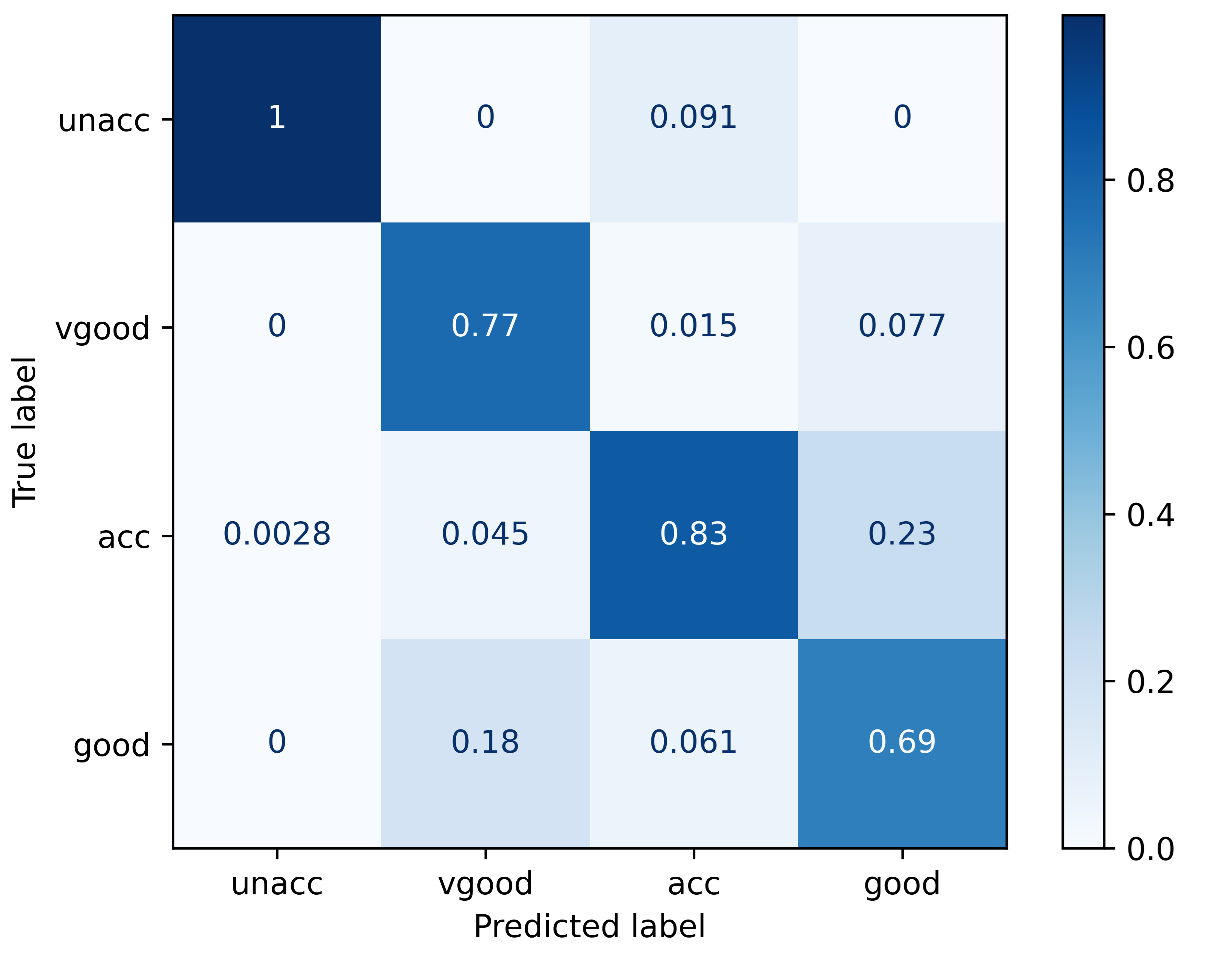}	
		\subcaption{}	
	\end{minipage}
	\hfill
	\begin{minipage}{7.5cm}
		\includegraphics[width=1.0\textwidth]{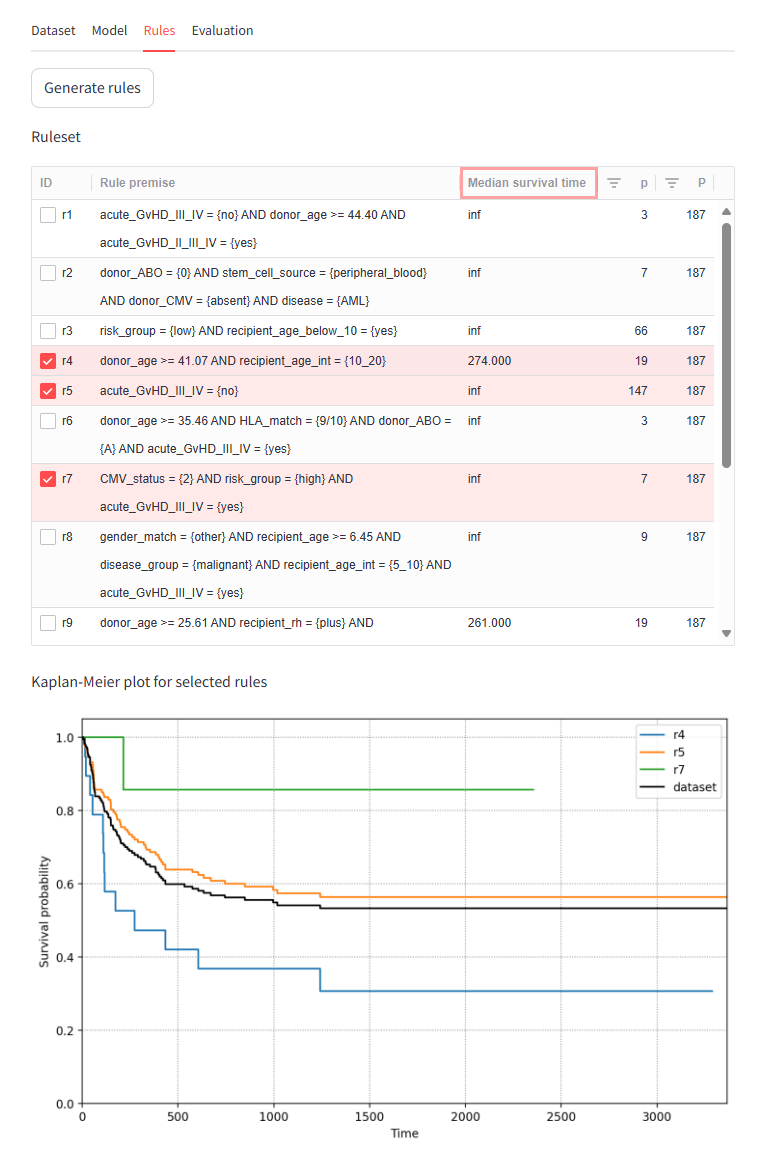}
		\subcaption{}			
	\end{minipage}
	\caption{RuleKit 2 usage examples. (a) Classification analysis of \textit{car} data set with Python package. (b) Visualization of the confusion matrix produced by the Python code. (c) Survival analysis of \textit{bone-marrow} data set using graphical user interface.}\label{fig:examples}
\end{figure}

\textbf{Algorithm scalability}
The cross validation analysis performed on over 120 data sets from classification, regression, and survival domains (Table~\ref{tab:results}) revealed RuleKit 2 to be 4--8 times faster than its predecessor. This was obtained without affecting the predictive power (balanced accuracy for classification, integrated Brier score for survival) or with improving it significantly (root relative squared error for regression; Wilcoxon signed-rank test $p$-value less than $10^{-5}$). The error reduction was thanks to the novel mean-based algorithm for inducing regression rules.

To demonstrate RuleKit 2 performance on large data, we selected three example data sets representing different kind of problems. As presented in Figure~\ref{fig:big-data}, the most impressive speed up was observed for classification problem where the total execution time of 10 cross validation folds dropped from 132 hours to 34 minutes (over $230\times$ improvement). The speed ups for regression and survival were, however, consistent with those obtained for smaller sets. The complexity (i.e., the number of rules) and the predictive power of classification and survival models remained unchanged at a significance level 0.01, as measured by the $t$-test. Nevertheless, the regression model was not only less complex but also exhibited significantly smaller error ($t$-test $p$-value less than $10^{-7}$) which confirmed the advantage of the new algorithm for regression rules over its predecessor.

All the data sets investigated within this study are publicly available from RuleKit webpage (\link{https://github.com/adaa-polsl/RuleKit}).   

\vspace{0.3cm}
\textbf{Python package}
As shown in Figure~\ref{fig:examples}a, rule-based data analysis in RuleKit-python package can be performed in just few lines of code. Importantly, thanks to the compilance of the presented utility with \textit{scikit-learn} library, the standard evaluation and visualisation mechanisms offered by \textit{scikit-learn} can be also straightforwardly employed (Figure~\ref{fig:examples}b). 

\vspace{0.3cm}
\textbf{Graphical user interface}
Friendly, web browser-based graphical user interface allows straightforward configuration of algorithms parameters and running the analyses (Figure~\ref{fig:examples}c). The visualization of survival function estimates corresponding to particular rules is a novel RuleKit 2 feature, not present in the previous release.

\section{Impact}

The exponential growth of data brought new challenges to the data mining field. Symbolic models are no exception, often expected to meet contradictory requirements. In particular, learning algorithms need to handle large, high dimensional, diversified datasets at the same time keeping the interpretability of resulting models. RuleKit 2 attempts to answer these needs. By successfully analyzing data sets consisting of hundreds thousands instances, it can contribute substantially to addressing new research challenges in the field, like human-supervised decision-making based on large data or knowledge discovery from large data.  

With more than 20,000 downloads and over 100 citations (articles \citep{gudys2020rulekit,sikora2019,wrobel2017}, Google Scholar, April 2025) RuleKit confirmed its usability in the community. The second, more scalable and intuitive version presented in this paper, has already been applied in a number of fields. In particular, RuleKit~2 has been used for research purposes by physicians from Department of Electrocardiology at Upper-Silesian Medical Centre to analyze cardiovascular data. The tool has been also employed in industry as a central component of RuleMiner, a commercial service for data mining and knowledge discovery (\url{https://ruleminer.ai/}) provided by Łukasiewicz Research Network –- Institute of Innovative Technologies EMAG~\citep{Sikora2024RuleMinerAI}. Finally, RuleKit 2 has found its application in academia -- it facilitates teaching Knowledge Discovery course for master students at Silesian University of Technology.

\section{Conclusions}

We show RuleKit 2, a substantially improved version of our package for rule-based analysis of classification, regression, and survival data. Presented tool includes three main novelties. First, the algorithmic developments significantly improved computational scalability enabling RuleKit 2 to be used for large data. Second, the Python package complying with \textit{scikit-learn} makes it convenient for data analysts. Finally, intuitive user interface facilitates decision-making and knowledge discovery for people with no programming skills like physicians.
These features and a fact that RuleKit 2 has already found applications in research, industry, and academia strengthens presented package as a useful data mining and knowledge discovery tool.










\bibliographystyle{elsarticle-num} 
\bibliography{rulekit2}



\end{document}